\documentclass[10pt,twocolumn,letterpaper]{article}

\usepackage{cvpr}      
\definecolor{cvprblue}{rgb}{0.21,0.49,0.74}
\usepackage[pagebackref,breaklinks,colorlinks,allcolors=cvprblue]{hyperref}

\def\paperID{} 
\def\confName{CVPR}
\def\confYear{2026}

\title{All Roads Lead to Rome: Incentivizing Divergent Thinking in Vision-Language Models}

\author{
    Xinyu Tian\textsuperscript{\rm 1} \qquad
    Shu Zou\textsuperscript{\rm 1,2} \qquad
    Zhaoyuan Yang\textsuperscript{\rm 3} \qquad
    Mengqi He\textsuperscript{\rm 1}\qquad
    Peter Tu\textsuperscript{\rm 3}\qquad
    Jing Zhang\textsuperscript{\rm 1}
    \\
    \textsuperscript{\rm 1}Australian National University \quad \textsuperscript{\rm 2}Shanghai AI Lab \quad 
    \textsuperscript{\rm 3}GE Research
    \\
    {\tt\small \textsuperscript{\rm 1}firstname.lastname@anu.edu.au, \textsuperscript{\rm 3}firstname.lastname@ge.com}
}

\begin{document}

\maketitle
\begin{abstract}
Recent studies have demonstrated that Reinforcement Learning (RL), notably Group Relative Policy Optimization (GRPO), can intrinsically elicit and enhance the reasoning capabilities of Vision-Language Models (VLMs). However, despite the promise, the underlying mechanisms that drive the effectiveness of RL models as well as their limitations remain underexplored. In this paper, we highlight a fundamental behavioral distinction between RL and base models, where the former engages in deeper yet narrow reasoning, while base models, despite less refined along individual path, exhibit broader and more diverse thinking patterns. Through further analysis of training dynamics, we show that GRPO is prone to diversity collapse, causing models to prematurely converge to a limited subset of reasoning strategies while discarding the majority of potential alternatives, leading to local optima and poor scalability. To address this, we propose Multi-Group Policy Optimization (MUPO), a simple yet effective approach designed to incentivize divergent thinking across multiple solutions, and demonstrate its effectiveness on established benchmarks. Project page: \href{https://xytian1008.github.io/MUPO/}{\textcolor[HTML]{d12d8a}{https://xytian1008.github.io/MUPO/}}.
\end{abstract}

\section{Introduction}

Since the advent of Large Language Models (LLMs)~\citep{touvron2023llama, Singh_2025, seed2025seed1}, reasoning has emerged as a critical capability for addressing complex tasks. Early approaches typically rely on prompt engineering~\citep{kojima2022large, zhou2022least}, which elicit chain-of-thought solutions, or on Supervised Fine-Tuning (SFT)~\citep{muennighoff2025s1, chen2024expanding}, where models are trained on high-quality trajectories to emulate human-like thinking. More recently, the rise of Reinforcement Learning (RL), notably Group Relative Policy Optimization (GRPO)~\citep{shao2024deepseekmath, schulman2017proximal}, facilitates self-reflection and verification, highlighting remarkable self-improving capabilities. Consequently, researchers have increasingly sought to integrate reasoning into Vision-Language Models (VLMs)~\citep{bai2025qwen2, yang2025qwen3, guo2025seed1}. By leveraging RL, VLMs acquire the ability to extract, analyze and reflect over visual information, achieving significant gains on both challenging logical visual questions~\citep{yang2025r1, huang2025vision} and traditional tasks~\citep{liu2025visual, shen2025vlm}. 

However, despite the promise, the underlying mechanisms driving the effectiveness of RL models, as well as their limitations, remain underexplored. Early doubts arise from LLMs, where \citep{yue2025does} observe that although RL models demonstrate higher accuracy, their performance ceilings remain constrained by the capabilities of base models, lacking the ability to expand novel reasoning patterns. More recently, in the context of VLMs, \citep{li2025think, xia2025visionary} notice that incorporating reasoning, in some cases, leads RL models to underperform base models. These findings lead us to ask: Do RL models truly outperform their base counterparts?

\begin{figure*}[t!] 
    \centering
\includegraphics[width=1\linewidth]{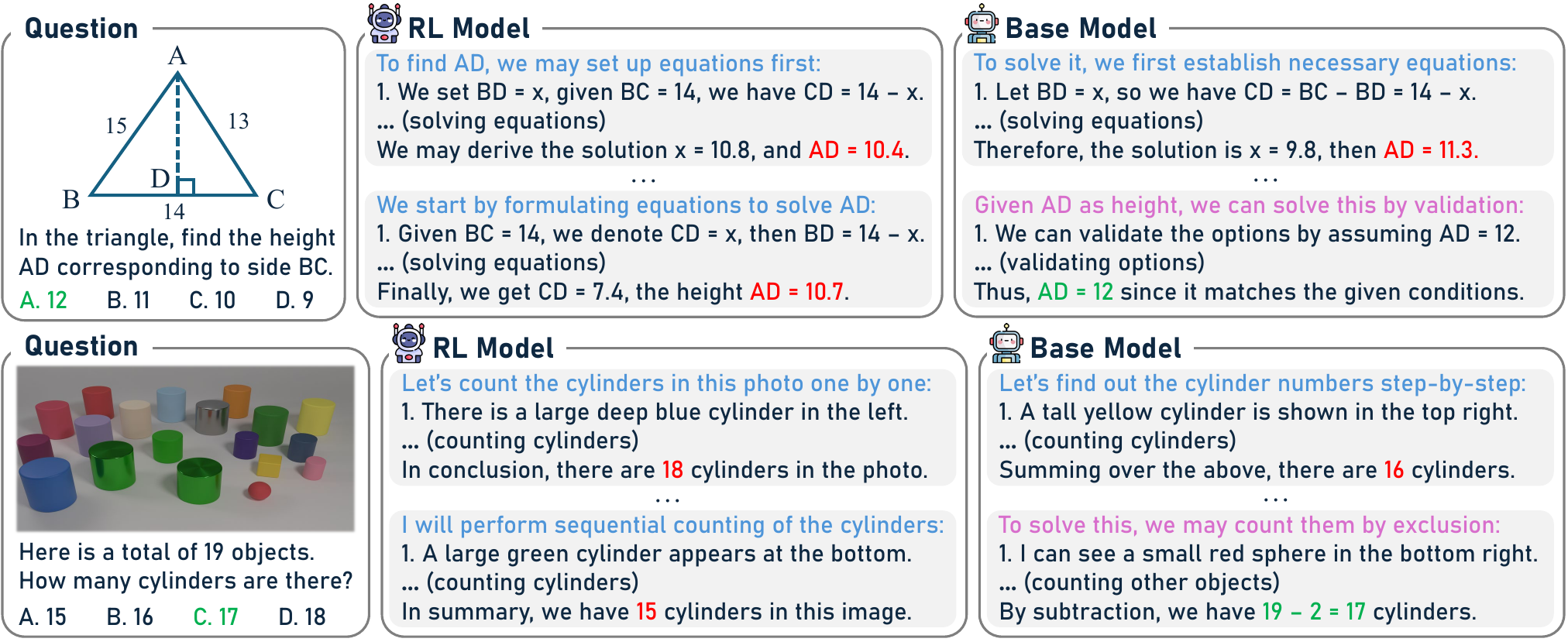}
    \vspace{-6mm}
    \caption{The failure cases of RL models. We use Vision-R1~\citep{huang2025vision} as a representative RL model, with its corresponding base model being Qwen2.5-VL-7B~\citep{bai2025qwen2}. The examples are selected from MathVerse~\citep{zhang2024mathverse} and MathVista~\citep{lu2023mathvista}, respectively. For each question, we set the sampling temperature to $1.0$ and generate multiple responses, each of which is displayed in a \textcolor[HTML]{797979}{gray} box. Main differences in the proposed reasoning strategies are annotated in \textcolor[HTML]{4E95D9}{blue} and \textcolor[HTML]{D86ECC}{pink}, while correct and incorrect answers are highlighted in \textcolor[HTML]{00B050}{green} and \textcolor[HTML]{FF0000}{red}, respectively.
    }
\label{fig:motivation}
\vspace{-3mm}
\end{figure*}

Motivated by this, we conduct a behavioral comparison between RL and base models on established benchmarks. Our results indicate that, when limited to a single attempt, RL models generally achieve higher accuracy than their base counterparts. However, when multiple samplings are permitted, base models are consistently capable of solving a broader number of problems. Notably, in failure cases where RL models are unable to handle despite multiple attempts, we find that base models often succeed by leveraging through diverse and alternative reasoning pathways that are not captured by RL models. For instance, as shown in Fig.~\ref{fig:motivation}, for geometry problems, RL models tend to rely exclusively on equation solving, which is prone to logical errors. In contrast, base models are capable of proposing simpler, verification-based strategies. Similarly, in object counting involving large quantities, RL models consistently adopt tedious sequential enumeration. Base models, however, are often able to employ efficient elimination strategies that reach the correct answer in significantly fewer steps.

These observations reveal a fundamental distinction between RL models and their vanilla base counterparts. During reasoning, RL models, despite demonstrating deeper deliberation, tend to be conservative, often adhering to a dominant strategy. In contrast, base models, although less refined along a single path, display divergent thinking, frequently exploring potential alternative solutions. Intuitively, the latter more closely aligns with human problem-solving, in which individuals, when given multiple attempts, often approach the same problem from varied perspectives, increasing their chances of success, especially on challenging tasks where common strategies may fail or be error-prone. Our further experiments reveal that models also exhibit similar patterns, where greater diversity in reasoning significantly enhances the probability of reaching correct answers.

The above findings indicate that models trained with RL, \ie, GRPO, appear to forgo the inherent divergent thinking exhibited by their base counterparts. To further investigate the cause of this shift, we examine the evolution of reasoning diversity throughout the training process. We observe that during the early training stage, diversity declines sharply to a negligible level, suggesting the model rapidly converges on a narrow set of strategies while discarding the vast majority of potential paths. As a result, the model concentrates on optimizing this limited subset for most of the training period. This collapse in diversity has two issues: 1) exploitation over exploration, where the model prioritizes a dominant strategy over exploring alternative modes, leading to local optima; 2) poor scalability, where convergent reasoning struggles to cover the broad spectrum of problems, thereby constraining test-time scaling capabilities.

Based on the above, a natural question is: can we preserve divergent thinking from base models during RL, enabling them to reason deeply about individual solutions while also maintaining a repertoire of diverse strategies? To address this, we propose Multi-Group Policy Optimization (MUPO), a drop-in replacement of GRPO to incentivize divergent reasoning. Inspired by the diversity collapse in GRPO, we partition model responses into multiple groups. Instead of computing advantages globally, MUPO performs localized advantage estimation within each group and introduces diversity reward to promote greater separation among groups. Intuitively, each group serves as a distinct realization of a reasoning strategy, and MUPO aims for the model to not only generate a range of diverse strategies but also to refine each effectively. Our result model, MUPO-Thinker-7B, demonstrates the ability to explore diverse reasoning paths in search of globally optimal solutions and achieves average gains of $2\sim 7\%$ over strong baselines on established benchmarks, setting a new state of the art.

\noindent In summary, our contributions are as follows:
\begin{itemize}
    \item We highlight a fundamental difference in reasoning behavior between RL models and base models: the former engage in deeper yet narrowly focused reasoning, whereas the latter, despite being less sophisticated, exhibit broader and more diverse thought patterns.
    \item We find that GRPO is prone to diversity collapse, causing the model to search within a narrow set of strategies while disregarding the majority of potential alternatives, leading to local optima and limited scaling capabilities.
    \item We propose MUPO, a straightforward yet effective policy algorithm designed to incentivize divergent thinking across multiple solutions, and demonstrate its effectiveness through comprehensive experimental validation.
\end{itemize}
\section{Related Work}
\begin{figure*}[t!] 
    \centering
\includegraphics[width=1\linewidth]{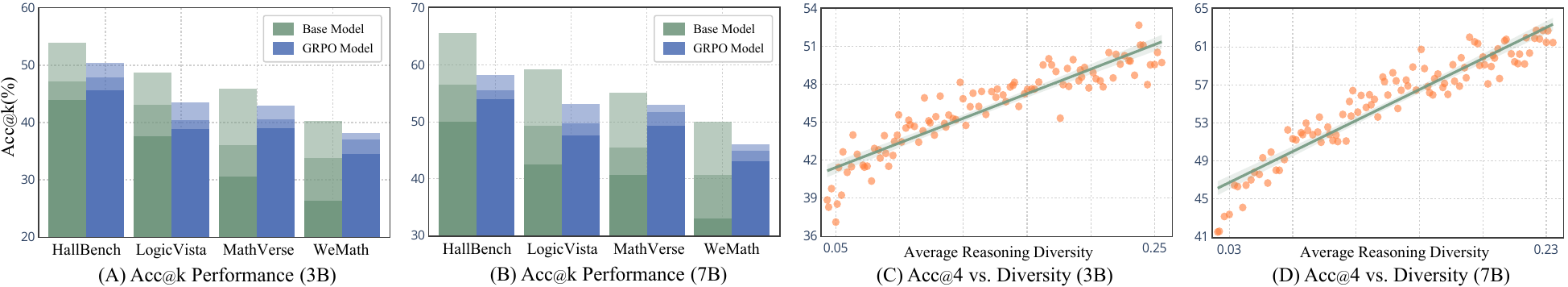}
    \vspace{-6mm}
    \caption{The impact of reasoning diversity on model performance. In (A) and (B), we report acc@k for both RL and base models on established benchmarks, with color intensity decreasing as $k=1,2,4$. In (C) and (D), we plot relationship between reasoning diversity and the corresponding acc@4 scores. Each point is based on a set of $4$ responses, and a regression line is fitted to capture the overall trend.
    }
\label{fig:accuracy}
\vspace{-3mm}
\end{figure*}
\noindent \textbf{Reasoning in VLMs}. Enhancing reasoning capabilities has become a critical objective in the pursuit of general intelligence. In the context of VLMs, this progression evolves from prompt engineering which elicits chain-of-thought reasoning~\citep{kojima2022large, zhou2022least, tian2024argue, tian2025black}, to SFT on high-quality, human-designed trajectories~\citep{muennighoff2025s1, cai2024internlm2}. More recently, the advent of RL, particularly GRPO~\citep{shao2024deepseekmath, schulman2017proximal}, has shifted the paradigm from passive distillation toward active self-improvement, empowering models to autonomously discover and refine optimal strategies. Consequently, a growing body of research explores reasoning via diverse RL strategies~\citep{peng2025skywork, tan2025reason, hu2025diversity}, dataset construction~\citep{yang2025r1, meng2025mm, tian2025identifying, yao2023training}, and reward shaping~\citep{wang2025perception, yang2025look, zou2025simlabel, yao2025simple}. Despite these advances, the mechanisms of RL models and their limitations remain underexplored. In this work, we show that while RL models exhibit higher accuracy, they tend to forgo the divergent thinking of base models, thereby limiting the potential to solve a broader range of problems.

\noindent \textbf{RL for Reasoning}. As researchers notice that sparse yet easily verifiable rewards, \eg, accuracy and format, yield unexpectedly strong performance in RL, subsequent studies have emerged aiming to advance GRPO along multiple dimensions. For instance, \citep{yu2025dapo, zhang2025gvpo} adopt a sampling-based perspective, prioritizing informative trajectories to enhance training efficiency. In parallel, \citep{tian2025more, jian2025look, zou2026unlocking} focus on visual features, promoting perceptually-aware reasoning through fine-grained reward design. Recently, another line of research~\citep{wang2025beyond, cheng2025reasoning, jiang2025risk} explores the integration of uncertainty via entropy mechanisms to encourage exploration. However, these approaches fail to foster reasoning across divergent strategies. In contrast, we propose MUPO, which enables models to reason with both depth and breadth, ultimately guiding them toward the discovery of optimal solutions.

\noindent \textbf{Test-Time Scaling}. Given the substantial expense of training, allocating extra compute at test time is becoming a cost-effective strategy to enhance performance~\citep{snell2024scaling, muennighoff2025s1}. In the case of VLMs, this manifests as allowing for more thinking budgets. Such scaling law can be categorized into two types: sequential and parallel. The former promotes deeper reasoning through patterns such as step-by-step generation, self-reflection, and verification~\citep{vl-rethinker, liu2025visual, he2025few}. The latter emphasizes divergent exploration by sampling multiple candidates and aggregating via self-consistency or verifiers~\citep{zheng2025parallel, wang2025visualprm}. However, existing RL algorithms predominantly focus on sequential thinking, \ie, refining along a single reasoning path, while neglecting alternative branches, which limits their scaling potentials. Our proposed method, MUPO, bridges this gap by integrating parallel thinking into RL, yielding significant gains in both accuracy and scalability.
\section{Exploring Divergent Thinking in VLMs}
In this section, we begin with a comparison between existing RL models and their base counterparts, focusing on diversity as a key axis. We examine how variations in reasoning diversity influence model performance. Building on this, we further investigate the dynamics of diversity throughout the GRPO training process, revealing a critical issue where models tend to optimize within a narrow subset of strategies while discarding other potential alternatives.

\begin{figure*}[t!] 
    \centering
\includegraphics[width=1\linewidth]{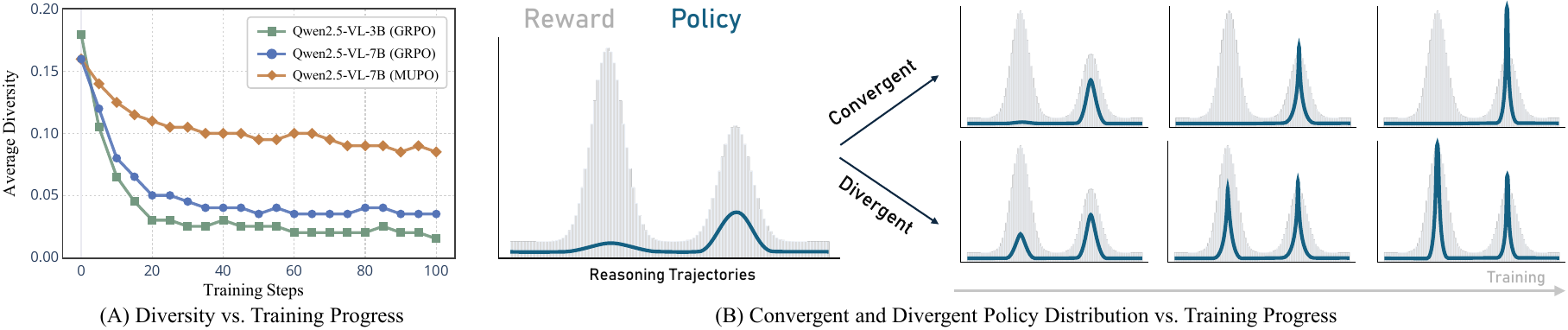}
    \vspace{-6mm}
    \caption{The diversity collapse of GRPO. In (A), we plot the evolution of reasoning diversity across training steps. In (B), we present an illustration of policy distribution over training to highlight the contrasting dynamics of convergent and divergent thinking. The gray region denotes rewards associated with different reasoning trajectories, while the blue curve indicates corresponding sampling probabilities.
    }
\label{fig:diversity}
\vspace{-3mm}
\end{figure*}

\subsection{Divergent vs Convergent}
\label{subsec:3.1}
To assess the impact of reasoning diversity to model performance, here we conduct a simple motivational experiment.

\noindent\textbf{Experimental Setup}. We consider models at two scales: 3B and 7B. For the 3B setting, we select VLAA-Thinker~\citep{chen2025sft} and LMM-R1~\citep{peng2025lmm}, while for the 7B setting, we consider Vision-R1~\citep{huang2025vision} and R1-OneVision~\citep{yang2025r1}, with Qwen2.5-VL-3B and Qwen2.5-VL-7B being the base models~\citep{bai2025qwen2}. All models are evaluated on a suite of reasoning-centric benchmarks, including MathVerse~\citep{zhang2024mathverse}, LogicVista~\citep{xiao2024logicvista}, WeMath~\citep{qiao2024we} and HallusionBench~\citep{guan2024hallusionbench}. The sampling temperature is set to $1.0$ to enable generation of multiple responses.

\noindent\textbf{Metric}. Beyond standard accuracy that only evaluates a single path, we introduce a more relaxed metric, acc@k, which is considered positive if at least one of $k$ sampled trajectories leads to the correct answer. We set $k=1$, $2$ and $4$ to assess the model's ability to reach the correct answer when given multiple attempts. To quantify reasoning diversity, we employ Qwen3-Embedding-0.6B~\citep{zhang2025qwen3} to encode the reasoning segments of generated responses and compute cosine distances to measure differences. The diversity across multiple trajectories is then calculated as the pairwise average.

\noindent\textbf{Results}. Fig.~\ref{fig:accuracy} displays the impact of reasoning diversity on models, from which we may derive following key insights.

\noindent\textbf{RL models dive depth, base models seek breadth}. As shown in Fig.~\ref{fig:accuracy} (A) and (B), when $k=1$, RL models markedly outperform their base counterparts, reflecting sophisticated reasoning along a single trajectory. However, as $k$ increases, a notable shift occurs: base models succeed in solving substantially more problems, while the gains of RL models remain marginal. This observation suggests that base models exhibit a greater capacity to generate effective alternative solutions in challenging cases, indicating their higher potential. Our qualitative analysis in Fig.~\ref{fig:motivation} further highlights this contrast, where RL models tend to adopt narrowly focused reasoning, yet base models display divergent reasoning, approaching problems from varied perspectives.

\begin{figure}[t!] 
    \centering
\includegraphics[width=1\linewidth]{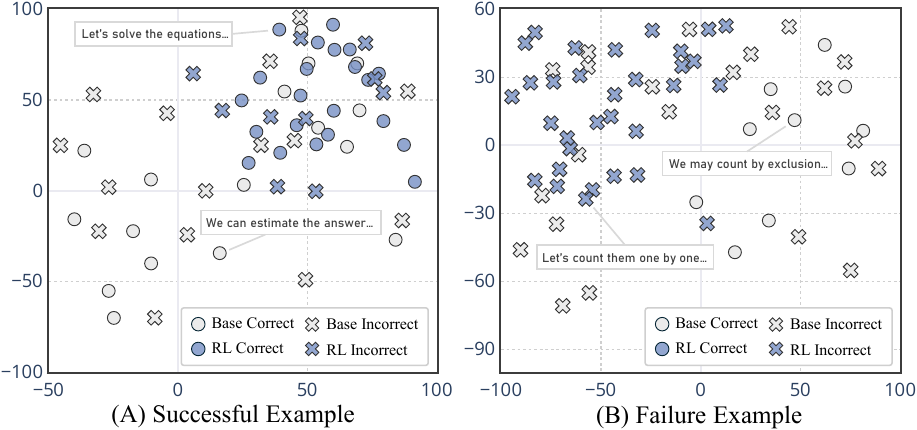}
    \vspace{-6mm}
    \caption{The t-SNE projection of reasoning embeddings. We analyze a successful case where RL models produce correct answers, and a failure case where they cannot despite multiple samplings.
    }
\label{fig:tsne}
\vspace{-3mm}
\end{figure}

\noindent\textbf{Divergent thinking increases the odds of success}. To further investigate the above distinctions, we perform multiple runs of base models and plot the relationship between reasoning diversity and corresponding acc@4 scores, as shown in Fig.~\ref{fig:accuracy} (C) and (D). We observe a strong positive correlation: as the reasoning diversity increases, acc@4 improves substantially. Intuitively, this indicates that tackling a problem through diverse paths rather than adhering to a single strategy facilitates the discovery of correct answers. This observation aligns with real-world problem-solving, where most tasks admit multiple viable solutions, and broader exploration often leads to more effective outcomes. Such findings also offer an explanation for the superior performance of base models over RL under parallel thinking.

To provide a deeper understanding, we visualize the t-SNE projection of reasoning embeddings for both RL and base models in Fig.~\ref{fig:tsne}. It reveals that RL reasoning embeddings are more densely clustered, while those of base models are more sparsely distributed. This structural difference explains higher pass rate of RL models in successful cases due to concentrated distribution. However, in failure cases, their narrow embedding region fails to cover any correct trajectory. In contrast, base models, benefiting from a wider and more diverse reasoning space, are more likely to retrieve correct solutions from alternative regions.

\subsection{Diversity Collapse}
\label{subsec:3.2}

\begin{figure*}[t!] 
    \centering
\includegraphics[width=1\linewidth]{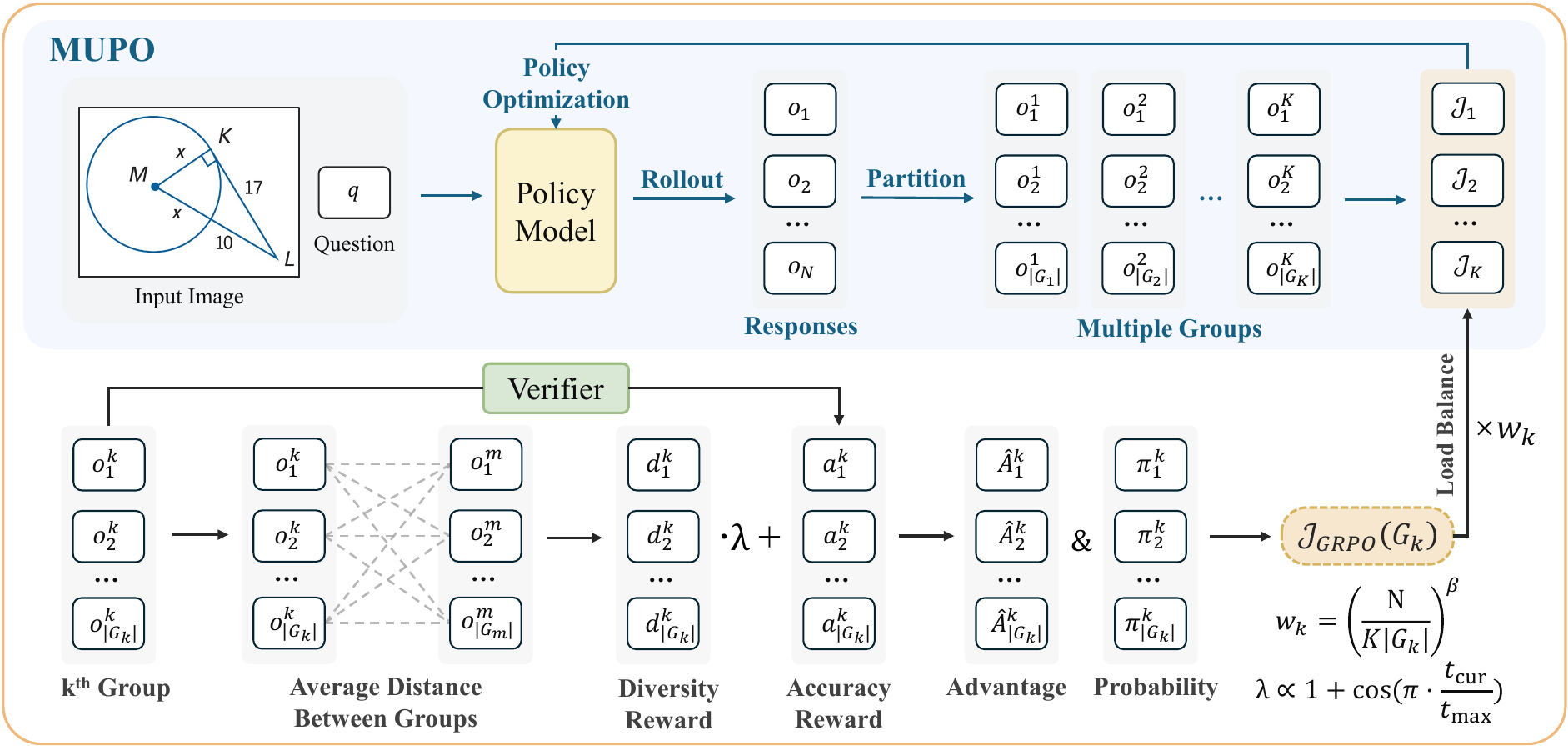}
    \vspace{-6mm}
    \caption{The overview of MUPO. The upper part illustrates the high-level pipeline, where responses are partitioned into multiple groups, and the overall optimization objective is formulated as a composition of multiple GRPO objectives, each corresponding to a group. In the lower part, we present the advantage computation for a group, in which we introduce diversity reward to encourage inter-group separation.
    }
\label{fig:method}
\vspace{-3mm}
\end{figure*}

The above findings indicate that RL, when applied to base models, diminishes inherent divergent thinking, steering them toward a single, specialized reasoning strategy. To gain deeper insights into this behavioral shift, we examine the evolution of diversity throughout the training process.

\noindent\textbf{Experimental Setup}. We adopt GRPO as the typical RL algorithm and consider two base models: Qwen2.5-VL-3B and Qwen2.5-VL-7B~\citep{bai2025qwen2}. The models are trained on ViRL39K~\citep{vl-rethinker}, a high-quality and comprehensive dataset, for around half an epoch. For each step, we generate $10$ responses per example and record diversity on validation set.

\noindent\textbf{GRPO narrows the mind before learning begins}. As shown in Fig.~\ref{fig:diversity}~(A), we observe that reasoning diversity drops sharply during the early training stage, \ie, first 20 steps, where the model has been exposed to only limited training data. This collapse suggests a premature convergence to a narrow set of strategies, resulting in the exclusion of most potential solution paths and dedicating the majority of training to refining only a small subset. Such phenomenon has two key issues. 1) Exploitation over exploration: the model favors unimodal optimization, neglecting alternative modes and thereby becoming susceptible to local optima; 2) Limited scalability: as discussed in Section~\ref{subsec:3.1}, this convergent reasoning fails to generalize across a broad range of questions, constraining the scaling capabilities.

In Fig.~\ref{fig:diversity}~(B), we provide a conceptual illustration of the training dynamics for convergent and divergent reasoning. Given that a problem may have multiple valid strategies, each corresponding to a distinct mode as shown in the figure, convergent training, \ie, GRPO, tends to select one mode early and progressively sharpens the distribution toward it. In contrast, divergent training, \ie, our expectation, encourages the model to explore and refine multiple modes, enabling the discovery of globally optimal solutions.
\section{The Proposed Method}
\subsection{Preliminary}
To introduce our approach, we first revisit the standard RL algorithm, notably Group Relative Policy Optimization (GRPO). In the multimodal setting, consider a dataset $\mathcal{D}$, where each example comprises a question $q$, a gold answer $y$, and corresponding visual input $I$. GRPO is designed to promote the generation of high-quality responses by contrasting multiple candidate outputs, rewarding superior ones while penalizing less effective alternatives. Given a policy model $\pi_{\theta}$, the objective is to maximize the following:

{\setlength{\abovedisplayskip}{-5pt}
\setlength{\belowdisplayskip}{6pt}
\begin{alignat}{2}
&\mathcal{J}_{\rm GRPO}(\theta) = 
\mathbb{E}_{(q, y, \mathcal{I})\sim \mathcal{D}, \{o_i\}^{\left|G\right|}_{i=1}\sim {\pi}_{\theta_{\rm old}}(\cdot \mid q, \mathcal{I})} \nonumber \\
& \ \ \left[\frac{1}{\left|G\right|}\sum_{i=1}^{\left|G\right|} \min\Bigl(r_{i}(\theta)\hat{A}_{i},
{\rm clip}\Bigl(r_{i}(\theta), 1 - \epsilon, 1 + \epsilon\Bigr)\hat{A}_{i}\Bigr)\right] \nonumber
\\ 
&{\rm with} \ r_{i}(\theta) = \frac{\pi_{\theta}(o_{i}\mid q, \mathcal{I})}{\pi_{\theta_{\rm old}}(o_{i}\mid q, \mathcal{I})},
\end{alignat}}

\noindent where $G$ represents the set of sequences $\{o_1, o_2, ..., o_{\left|G\right|}\}$ sampled from the old policy $\pi_{\rm old}$, $\epsilon$ controls the clipping bounds to maintain on-policy training. It is optional to impose a KL constraint toward the reference policy for stable optimization. The estimated token-level advantage $\hat{A}_{i}$ is derived by broadcasting the normalized sequence-level reward $R_{i}$ across all token positions, which is defined as follows:

{\setlength{\abovedisplayskip}{-6pt}
\setlength{\belowdisplayskip}{4pt}
\begin{align}
    \hat{A}_{i} = \frac{R_i - {\rm mean}\left(\mathbf{R}\right)}{{\rm std}\left(\mathbf{R}\right)}, \quad i=1,\cdots,\left|G\right|,
\end{align}}

\noindent where $\mathbf{R}=\{R_1, R_2, ..., R_{\left|G\right|}\}$ indicates the reward of the sequence group. A common choice for reward design is the verifiable reward, which provides direct feedback based on metrics such as accuracy and format consistency.

\subsection{Multi-Group Policy Optimization}
Despite the promise of GRPO, our findings in Section~\ref{subsec:3.2} reveal that it is prone to diversity collapse, where models prematurely converge to a subset of reasoning paths while abandoning alternative strategies, leading to local optima. To address this, we propose Multi-Group Policy Optimization (MUPO), a simple yet effective drop-in replacement of GRPO to incentivize divergent thinking. Inspired by the intra-group diversity collapse, MUPO partitions responses into multiple groups. Unlike GRPO which computes advantages globally, MUPO performs localized advantage estimation within each group, and introduces a diversity reward to promote inter-group separation. Intuitively, each group serves as a distinct realization of a reasoning strategy, and MUPO aims to maintain multiple modes while refining each of them, thereby achieving both breadth and depth. The overview of our approach has been presented in Fig.~\ref{fig:method}.

\noindent\textbf{Multi-Group Objective}. Given $N$ sampled responses at each step, we partition them into $K$ groups $\{G_k\}_{k=1}^{K}$ based on the embedding space as described in Section~\ref{subsec:3.1}. Specifically, we apply constrained clustering~\citep{bradley2000constrained} to group responses with similar trajectories, while enforcing minimum group size $G_{\rm min}$ for reliable advantage estimation. This enables each group to capture a distinct reasoning mode. The objective of MUPO is to maximize the following:

{\setlength{\abovedisplayskip}{-8pt}
\setlength{\belowdisplayskip}{6pt}
\begin{alignat}{2}
&\mathcal{J}_{\rm MUPO}(\theta) = \mathbb{E}_{(q, y, \mathcal{I})\sim \mathcal{D}, \{o_i\}^{N}_{i=1}\sim {\pi}_{\theta_{\rm old}}(\cdot \mid q, \mathcal{I})}
\nonumber \\ 
&\hspace{-1mm}\Biggl[\sum_{k=1}^{K}\frac{w_{k}}{\left|G_{k}\right|}\underbrace{\sum_{i=1}^{\left|G_{k}\right|} \min\Bigl(r_{i}(\theta)\hat{A}_{i}^{k},
{\rm clip}\Bigl(r_{i}(\theta), 1 - \epsilon, 1 + \epsilon\Bigr)\hat{A}_{i}^{k}\Bigr)}_\text{$\mathcal{J}_{\rm GRPO}$}\Biggr] \nonumber \\ 
&{\rm with} \ r_{i}(\theta) = \frac{\pi_{\theta}(o_{i}^{k}\mid q, \mathcal{I})}{\pi_{\theta_{\rm old}}(o_{i}^{k}\mid q, \mathcal{I})}, \ \ w_{k} = \left(\frac{N}{K\left|G_{k}\right|}\right)^{\beta}.
\end{alignat}}

\noindent Essentially, MUPO can be regarded as a composition of multiple GRPO objectives, enabling the search for optima from diverse modes. $w_{k}$ is a load-balance scaler controlled by sensitivity exponent $\beta$, which modulates the contribution of each group to the overall objective, preventing larger groups from dominating the optimization process. The advantage $\hat{A}_{i}^{k}$ is estimated locally within each group to ensure the refinement of each mode proceeds independently:

{\setlength{\abovedisplayskip}{-5pt}
\setlength{\belowdisplayskip}{6pt}
\begin{align}
    \hat{A}_{i}^{k} = \frac{R_{i}^{k} - {\rm mean}\left(\mathbf{R}\right)}{{\rm std}\left(\mathbf{R}\right)}, \quad i=1,\cdots,\left|G_{k}\right|.
\end{align}}

\noindent\textbf{Diversity Reward}. To encourage models to explore various reasoning strategies, we introduce a diversity reward that increases the separation between groups. Specifically, for a given trajectory $o_{i}^{k}$ from $G_{k}$, we compute its diversity reward as the average distance between its reasoning embedding and those of responses from all other groups:

{\setlength{\abovedisplayskip}{-6pt}
\setlength{\belowdisplayskip}{4pt}
\begin{align}
    R_{\rm div} = \frac{1}{N-\left|G_{k}\right|}\sum_{\substack{m=1 \\ m\ne k}}^{K}\sum_{j=1}^{\left|G_{m}\right|} d(o_{i}^{k}, o_{j}^{m}),
\end{align}}

\noindent where $o_{j}^{m}$ indicates the trajectory from $G_{m}$, and $d(\cdot,\cdot)$ denotes the cosine distance between the reasoning embeddings of two responses. This ensures that responses within a group that are more distant from other groups receive a higher advantage. The final reward for $o_{i}^{k}$ is computed as:

{\setlength{\abovedisplayskip}{-6pt}
\setlength{\belowdisplayskip}{5pt}
\begin{align}
    R_{i}^{k} = R_{\rm acc} + R_{\rm fmt} + \lambda \cdot \mathbf{1}\left[R_{\rm acc} = 1\right]\cdot R_{\rm div}.
\end{align}}

\noindent $R_{\rm acc}$ and $R_{\rm fmt}$ denote accuracy and format reward, respectively. We impose an accuracy condition on the diversity reward to prevent reward hacking, where models pursue diverse outputs at the expense of correctness. $\lambda$ is the weight of the diversity reward and is annealed over the current training step $t_{\rm cur}$ according to the cosine schedule:

{\setlength{\abovedisplayskip}{-6pt}
\setlength{\belowdisplayskip}{5pt}
\begin{align}
    \lambda = \lambda_{\rm min} + \frac{\lambda_{\rm max} - \lambda_{\rm min}}{2}  \left(1+\cos\left(\pi \cdot \frac{t_{\rm cur}}{t_{\rm max}}\right)\right),
\end{align}}

\noindent where $t_{\rm max}$ is the total number of training steps, $\lambda_{\rm max}$ and $\lambda_{\rm min}$ specify the desired initial and final values of $\lambda$ with a smooth and monotonic decay. This design encourages broad exploration of diverse reasoning strategies in the early training, while gradually shifting the focus toward identifying globally optimal solutions in later stages.

\section{Experiment and Results}
\noindent\textbf{Implementation Details}. Similar to settings in Section~\ref{subsec:3.2}, we train all models on ViRL39K~\citep{vl-rethinker} for 2 epochs with a learning rate of $1e^{-6}$. A random subset of the dataset is used in ablation study for efficiency. We select Qwen2.5-VL~\citep{bai2025qwen2} at 3B and 7B parameter scales as base models. For group reward computation, we generate $N=15$ responses per example with a sampling temperature of $1.0$. For MUPO, unless otherwise specified, we partition responses into $K=3$ groups with a minimum size of $G_{\rm min}=3$. We set the load-balance exponent $\beta=1$, and the initial and final values of diversity reward weight $\lambda_{\rm max}=0.4$ and $\lambda_{\rm min}=0.1$. 

\noindent\textbf{Benchmarks}. We denote models trained with our method as MUPO-Thinker and evaluate on nine reasoning benchmarks spanning various task types: mathematical benchmarks including MathVerse~\citep{zhang2024mathverse}, MathVista~\citep{lu2023mathvista}, MathVision~\citep{wang2024measuring}, LogicVista~\citep{xiao2024logicvista}, WeMath~\citep{qiao2024we}, and Geometry3K~\citep{lu2021inter}, as well as general-purpose ones encompassing MMStar~\citep{chen2024we}, HallusionBench~\citep{guan2024hallusionbench} and MMVet~\citep{yu2023mm}.

\noindent\textbf{Baseline Methods}. To verify the effectiveness of our approach, we compare MUPO-Thinker against existing strong baselines, including InternVL2.5~\citep{chen2024expanding}, R1-OneVision~\citep{yang2025r1}, VLAA-Thinker~\citep{chen2025sft}, Vision-R1~\citep{huang2025vision}, VLM-R1~\citep{shen2025vlm} and LMM-R1~\citep{peng2025lmm}, spanning both 3B and 7B models. We also report scores of proprietary models such as GPT-5-Thinking~\citep{Singh_2025} and Gemini-2.5-Pro~\citep{comanici2025gemini} for reference. The baseline results are primarily cited from the corresponding papers, secondarily from the OpenCompass leaderboard, and reproduced when neither source is available.

\noindent\textbf{Metrics}. To evaluate both the depth and breadth of reasoning, we report acc@1 and acc@4 as defined in Section~\ref{subsec:3.1}. For the former, we employ greedy decoding, which yields the most confident responses. For the latter, we generate 4 candidates per example with a sampling temperature of $1.0$, thereby assessing the model's test-time scaling capabilities. Please refer to Appendix~A for more configuration details.

\begin{table*}[t]
\footnotesize
\centering
\setlength{\tabcolsep}{0.95mm}
\setlength\heavyrulewidth{0.25ex}
\renewcommand{\arraystretch}{1.1}
\begin{adjustbox}{width=\textwidth,center}
\begin{tabular}{@{}lcccccccccccccc@{}}
\toprule
\multicolumn{1}{c|}{\multirow{2}{*}{Models}} & \multicolumn{2}{c|}{MathVerse~\citep{zhang2024mathverse}}                     & \multicolumn{2}{c|}{MathVista~\citep{lu2023mathvista}}                     & \multicolumn{2}{c|}{MathVision~\citep{wang2024measuring}}                    & \multicolumn{2}{c|}{LogicVista~\citep{xiao2024logicvista}}                    & \multicolumn{2}{c|}{WeMath~\citep{qiao2024we}}                        & \multicolumn{2}{c|}{Geometry3k~\citep{lu2021inter}}                         & \multicolumn{2}{c}{Average}   \\ \cmidrule(){2-15} 
\multicolumn{1}{c|}{}                        & Acc@1         & \multicolumn{1}{c|}{Acc@4}         & Acc@1         & \multicolumn{1}{c|}{Acc@4}         & Acc@1         & \multicolumn{1}{c|}{Acc@4}         & Acc@1         & \multicolumn{1}{c|}{Acc@4}         & Acc@1         & \multicolumn{1}{c|}{Acc@4}         & Acc@1         & \multicolumn{1}{c|}{Acc@4}         & Acc@1         & Acc@4         \\ \midrule
\multicolumn{15}{c}{\textit{Closed-Source Models}}                                                                                                                                                                                                                                                                                                                                                          \\ \midrule
\multicolumn{1}{l|}{GPT-5-Thinking*~\citep{wang2025perception}}          & 81.2          & \multicolumn{1}{c|}{85.5}          & 81.9          & \multicolumn{1}{c|}{86.1}          & 72.0          & \multicolumn{1}{c|}{79.2}          & 70.0          & \multicolumn{1}{c|}{81.5}          & 71.1          & \multicolumn{1}{c|}{78.4}          & 79.9          & \multicolumn{1}{c|}{84.3}          & 76.1          & 82.5          \\
\multicolumn{1}{l|}{Gemini-2.5-Pro*~\citep{comanici2025gemini}}          & 76.9          & \multicolumn{1}{c|}{79.3}          & 80.9          & \multicolumn{1}{c|}{85.2}          & 69.1          & \multicolumn{1}{c|}{72.5}          & 73.8          & \multicolumn{1}{c|}{76.4}          & 78.0          & \multicolumn{1}{c|}{82.7}          & 77.2          & \multicolumn{1}{c|}{80.1}          & 76.0          & 79.4          \\ \midrule
\multicolumn{15}{c}{\textit{Open-Source Models}}                                                                                                                                                                                                                                                                                                                                                           \\ \midrule
\multicolumn{1}{l|}{Qwen2.5-VL-7B~\citep{bai2025qwen2}}           & 40.7          & \multicolumn{1}{c|}{55.2}          & 62.3          & \multicolumn{1}{c|}{\underline{78.5}}          & 23.2          & \multicolumn{1}{c|}{\textbf{41.6}} & 42.6          & \multicolumn{1}{c|}{\underline{59.3}}          & 33.1          & \multicolumn{1}{c|}{\textbf{50.1}} & 38.5          & \multicolumn{1}{c|}{54.4}          & 40.1          & \underline{56.5}          \\
\multicolumn{1}{l|}{InternVL2.5-8B*~\citep{chen2024expanding}}          & 34.5          & \multicolumn{1}{c|}{42.3}          & 68.2          & \multicolumn{1}{c|}{72.8}          & 25.6          & \multicolumn{1}{c|}{29.4}          & 38.3          & \multicolumn{1}{c|}{44.7}          & 38.6          & \multicolumn{1}{c|}{43.5}          & 44.8          & \multicolumn{1}{c|}{48.0}          & 41.7          & 46.8          \\
\multicolumn{1}{l|}{R1-OneVision-7B~\citep{yang2025r1}}         & 46.4          & \multicolumn{1}{c|}{49.9}          & 64.1          & \multicolumn{1}{c|}{69.4}          & \underline{29.9}          & \multicolumn{1}{c|}{34.1}          & 45.6          & \multicolumn{1}{c|}{52.5}          & \textbf{44.6} & \multicolumn{1}{c|}{47.9}          & 46.1          & \multicolumn{1}{c|}{50.4}          & 46.1          & 50.7          \\
\multicolumn{1}{l|}{VLAA-Thinker-7B~\citep{chen2025sft}}         & 48.2          & \multicolumn{1}{c|}{51.6}          & 68.0          & \multicolumn{1}{c|}{70.8}          & 26.4          & \multicolumn{1}{c|}{30.3}          & 48.5          & \multicolumn{1}{c|}{54.5}          & 41.5          & \multicolumn{1}{c|}{46.8}          & \underline{50.6}          & \multicolumn{1}{c|}{\underline{55.2}}          & 47.2          & 51.5          \\
\multicolumn{1}{l|}{Vision-R1-7B~\citep{huang2025vision}}            & \underline{52.4}          & \multicolumn{1}{c|}{\underline{56.1}}          & \underline{73.5}          & \multicolumn{1}{c|}{75.6}          & 28.2          & \multicolumn{1}{c|}{32.9}          & \underline{49.7}          & \multicolumn{1}{c|}{53.8}          & 41.6          & \multicolumn{1}{c|}{44.3}          & 49.0          & \multicolumn{1}{c|}{54.1}          & \underline{49.1}          & 52.8          \\ \midrule
\multicolumn{15}{c}{\textit{Our Models}}                                                                                                                                                                                                                                                                                                                                                                   \\ \midrule
\multicolumn{1}{l|}{MUPO-Thinker-3B}         & 41.3          & \multicolumn{1}{c|}{50.8}          & 64.3          & \multicolumn{1}{c|}{72.5}          & 27.8          & \multicolumn{1}{c|}{35.4}          & 42.8          & \multicolumn{1}{c|}{50.3}          & 36.5          & \multicolumn{1}{c|}{46.8}          & 45.1          & \multicolumn{1}{c|}{52.9}          & 43.0          & 51.5          \\
\rowcolor[HTML]{EDEDED} \multicolumn{1}{l|}{MUPO-Thinker-7B}         & \textbf{53.9} & \multicolumn{1}{c|}{\textbf{61.7}} & \textbf{77.9} & \multicolumn{1}{c|}{\textbf{82.4}} & \textbf{31.3} & \multicolumn{1}{c|}{\underline{39.7}}          & \textbf{50.6} & \multicolumn{1}{c|}{\textbf{61.5}} & \underline{44.1}          & \multicolumn{1}{c|}{\underline{48.6}}          & \textbf{52.1} & \multicolumn{1}{c|}{\textbf{59.1}} & \textbf{51.6} & \textbf{58.8} \\ \bottomrule
\end{tabular}
\end{adjustbox}
\vspace{-2mm}
\caption{The evaluation of our method on mathematical benchmarks. We report scores of both open-source and proprietary VLMs. The best and second best results are \textbf{bolded} and \underline{underlined}, respectively. * indicates the results sourced from the OpenCompass leaderboard.}
\label{tab:math}
\vspace{-4mm}
\end{table*}

\subsection{Main Results}

\begin{table}[t]
\footnotesize
\centering
\setlength{\tabcolsep}{0.4mm}
\setlength\heavyrulewidth{0.25ex}
\renewcommand{\arraystretch}{1.2}
\begin{adjustbox}{width=\columnwidth,center}
\begin{tabular}{@{}l|cc|cc|cc|cc@{}}
\toprule
\multicolumn{1}{c|}{\multirow{2}{*}{Models}} & \multicolumn{2}{c|}{MMStar~\citep{chen2024we}}   & \multicolumn{2}{c|}{HallBench~\citep{guan2024hallusionbench}} & \multicolumn{2}{c|}{MMVet~\citep{yu2023mm}}    & \multicolumn{2}{c}{Average}   \\ \cmidrule(){2-9} 
\multicolumn{1}{c|}{}                        & Acc@1         & Acc@4         & Acc@1          & Acc@4         & Acc@1         & Acc@4         & Acc@1         & Acc@4         \\ \midrule
QwenVL~\citep{bai2025qwen2} \quad                                    & 59.2          & {\ul 74.8}    & 50.0           & \textbf{65.6} & 64.8          & {\ul 74.1}    & 58.0          & {\ul 71.5}    \\
InternVL~\citep{chen2024expanding}                                     & 63.2          & 70.6          & 49.0           & 58.3          & 62.8          & 69.0          & 58.3          & 66.0          \\
R1-OV~\citep{yang2025r1}                                        & 64.7          & 67.5          & 52.5           & 57.6          & 65.2          & 67.3          & 60.8          & 64.1          \\
VLAA~\citep{chen2025sft}                                         & 66.1          & 69.1          & 54.7           & 56.9          & {\ul 70.0}    & 72.7          & {\ul 63.6}    & 66.2          \\
V-R1~\citep{huang2025vision}                                         & {\ul 66.3}    & 68.9          & {\ul 55.4}     & 59.0          & 68.2          & 70.5          & 63.3          & 66.1          \\
\rowcolor[HTML]{EDEDED} MUPO \quad                                        & \textbf{68.7} & \textbf{75.4} & \textbf{57.5}  & {\ul 63.6}    & \textbf{70.6} & \textbf{78.2} & \textbf{65.6} & \textbf{72.4} \\ \bottomrule
\end{tabular}
\end{adjustbox}
\vspace{-2mm}
\caption{The results of MUPO on general-purpose benchmarks. All models considered in this evaluation are of the 7B scale.}
\label{tab:general}
\vspace{-3mm}
\end{table}

\noindent\textbf{MUPO establishes a new state of the art}. As shown in Table~\ref{tab:math} and \ref{tab:general}, MUPO-Thinker-7B achieves an average acc@1 improvement of $2.5\%$ ($49.1\% \rightarrow 51.6\%$) on mathematical benchmarks and $2.3\%$ ($63.3\% \rightarrow 65.6\%$) on general-purpose benchmarks over previous best results. Similarly, in Table~\ref{tab:3b}, MUPO-Thinker-3B surpasses existing strong baselines of the same scale with notable gains of $2.0\%$ ($41.5\% \rightarrow 43.5\%$) and $2.4\%$ ($55.4\% \rightarrow 57.8\%$) on two types of benchmarks, respectively. This suggests that incorporating divergent thinking enables models to discover globally superior solutions and demonstrate more sophisticated reasoning along individual path, highlighting the importance of reasoning diversity for effective RL training.

\begin{figure}[t!] 
    \centering
\includegraphics[width=1\linewidth]{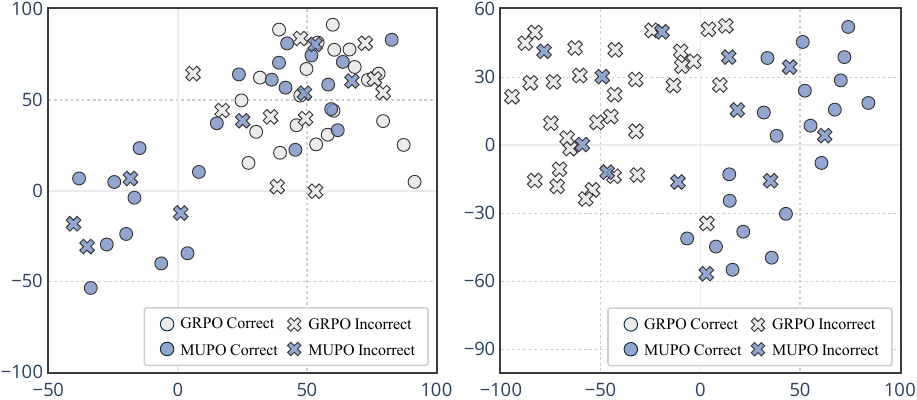}
    \vspace{-6mm}
    \caption{The t-SNE projection of our reasoning embeddings. The selected examples here correspond to the ones presented in Fig.~\ref{fig:tsne}.
    }
\label{fig:our_tsne}
\vspace{-4mm}
\end{figure}

\noindent\textbf{MUPO exhibits stronger test-time scaling capabilities}. As shown in Table~\ref{tab:math} and \ref{tab:general}, MUPO-Thinker-7B significantly outperforms existing strong RL models in acc@4 by $6.0\%$ ($52.8\% \rightarrow 58.8\%$) on mathematical benchmarks and $6.2\%$ ($66.2\% \rightarrow 72.4\%$) on general-purpose benchmarks, while surpassing the base model of the same scale. In Table~\ref{tab:3b}, MUPO-Thinker-3B also shows a substantial performance gap over recent baseline models, achieving an average gain of $5.9\%$ ($50.1\% \rightarrow 56.0\%$). Moreover, this scalability improvement enables MUPO-Thinker-3B to attain performance comparable to several strong 7B baselines. These results demonstrate that our approach successfully integrates the complementary strengths of RL and base models, significantly raising the upper bound of their capabilities.

\subsection{Further Discussion}


\begin{table}[t]
\footnotesize
\centering
\setlength{\tabcolsep}{1mm}
\setlength\heavyrulewidth{0.25ex}
\renewcommand{\arraystretch}{1.2}
\begin{adjustbox}{width=\columnwidth,center}
\begin{tabular}{@{}l|c|cc|cc|cc@{}}
\toprule
\multicolumn{1}{c|}{\multirow{2}{*}{Models}} & \multirow{2}{*}{Params} & \multicolumn{2}{c|}{Mathematics}     & \multicolumn{2}{c|}{General}  & \multicolumn{2}{c}{Average}   \\ \cmidrule(l){3-8} 
\multicolumn{1}{c|}{}                        &                         & Acc@1         & Acc@4         & Acc@1         & Acc@4         & Acc@1         & Acc@4         \\ \midrule
QwenVL~\citep{bai2025qwen2}                                       & 3B                      & 33.4          & {\ul 47.0}    & 51.3          & {\ul 62.3}    & 40.0          & {\ul 52.1}    \\
InternVL~\citep{chen2024expanding}                                     & 2B                      & 26.5          & 34.4          & 53.1          & 58.9          & 35.3          & 42.6          \\
VLM-R1~\citep{shen2025vlm}                                       & 4B                      & 37.8          & 42.3          & 53.9          & 57.7          & 43.2          & 47.4          \\
VLAA~\citep{chen2025sft}                                         & 3B                      & 39.5          & 43.3          & 55.1          & 59.3          & 44.7          & 48.6         \\
LMM-R1~\citep{peng2025lmm}                                       & 4B                      & {\ul 41.5}    & 45.5          & {\ul 55.4}    & 59.2          & {\ul 46.1}    & 50.1          \\
\rowcolor[HTML]{EDEDED} MUPO                                         & 3B                      & \textbf{43.5} & \textbf{51.5} & \textbf{57.8} & \textbf{65.2} & \textbf{48.3} & \textbf{56.0} \\ \bottomrule
\end{tabular}
\end{adjustbox}
\vspace{-2mm}
\caption{The comparison of MUPO with baselines at 3B scale. We report average scores of mathematical and general benchmarks.}
\label{tab:3b}
\vspace{-4mm}
\end{table}

\begin{figure*}[t!] 
    \centering
\includegraphics[width=1\linewidth]{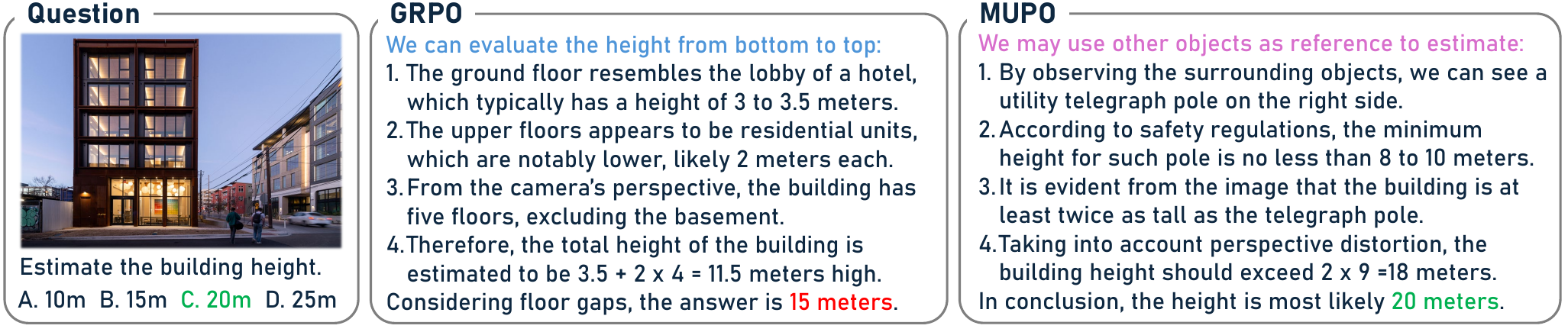}
    \vspace{-6mm}
    \caption{The qualitative analysis. We consider Vision-R1~\citep{huang2025vision} and MUPO-Thinker as typical GRPO and MUPO models and evaluate on an example from MMStar~\citep{chen2024we} with greedy decoding. We reorganize responses with numbering and omit special tags for better visualization.
    }
\label{fig:example}
\vspace{-3mm}
\end{figure*}

\begin{figure}[t!] 
    \centering
\includegraphics[width=1\linewidth]{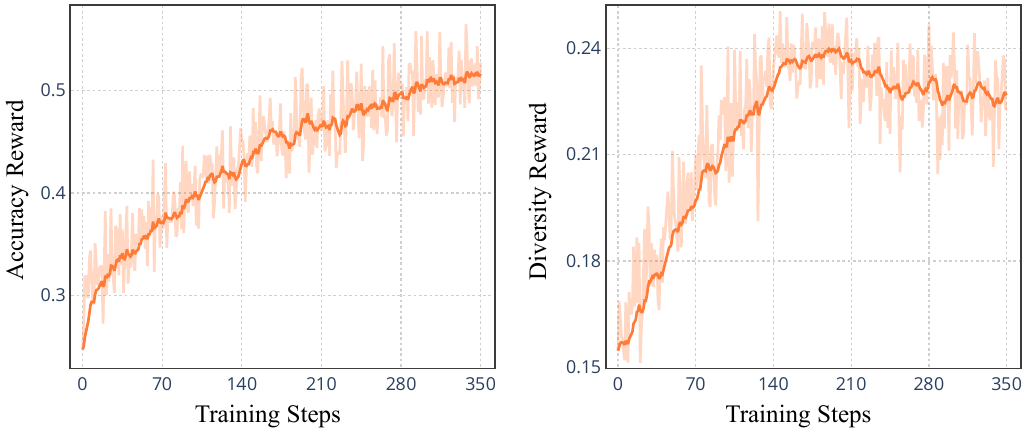}
    \vspace{-6mm}
    \caption{The learning curves of accuracy and diversity reward. We use exponential moving average (EMA) for smoothing.
    }
    \vspace{-2mm}
\label{fig:reward}
\end{figure}

\noindent\textbf{MUPO approaches problems with diverse strategies}. To verify whether MUPO indeed induces divergent thinking, we visualize the distribution of reasoning embeddings produced by our model, as shown in Fig.~\ref{fig:our_tsne}. In contrast to GRPO, which tends to sample from a narrow region, MUPO exhibits a broad, multimodal structure, with each mode corresponding to a distinct solution strategy. This proves particularly advantageous in the failure case on the right, where GRPO fails to handle the problem, while MUPO successfully learns the correct reasoning from alternative modes, providing a clear explanation for its superior performance.

\noindent\textbf{MUPO is capable of discovering better solutions}. To further understand the benefits of divergent training in MUPO, we conduct a qualitative analysis in Fig.~\ref{fig:example}. In this spatial reasoning example, where the task is to estimate the height of a building, GRPO adopts a rigid layer-by-layer estimation, which is prone to cumulative errors and ultimately fails. In contrast, our model leverages surrounding reference objects to precisely locate the building height range. This illustrates that MUPO enables models to learn a broader set of reasoning strategies and to generate smarter solutions when faced with various problem types.

\begin{table}[t]
\footnotesize
\centering
\setlength{\tabcolsep}{1mm}
\setlength\heavyrulewidth{0.25ex}
\renewcommand{\arraystretch}{1.1}
\begin{tabular}{@{}ccccccc@{}}
\toprule
\ K & MathVerse     & MathVista     & MathVision    & MMStar        & HallBench     & Average       \\ \midrule
\ 1 & 46.9          & 69.1          & 24.1          & 64.8          & 54.7          & 51.9          \\
\ 2 & 49.4          & 72.3          & 28.0          & 65.2          & \textbf{56.7} & 54.3          \\
\ 3 & \textbf{51.2} & 74.1          & 29.3          & \textbf{65.8} & 56.5          & \textbf{55.4} \\
\ 4 & 50.9          & 74.6          & \textbf{29.4} & 65.1          & 56.3          & 55.3          \\
\ 5 & 50.6          & \textbf{74.8} & 29.1          & 64.5          & 55.9          & 55.0          \\ \bottomrule
\end{tabular}
\vspace{-2mm}
\caption{The benchmark accuracy varying number of groups $K$.}
\label{tab:ablation_k}
\vspace{-4mm}
\end{table}

\noindent\textbf{MUPO learns from exploration to exploitation}. In Fig.~\ref{fig:reward}, we plot the learning curves of accuracy and diversity reward to further analyze the training dynamics of MUPO. As accuracy steadily improves, the diversity reward exhibits a distinct rise-fall-plateau trend. The initial rise indicates increasing distances between response groups, suggesting the model is actively exploring diverse modes. The slight decline in the middle is attributed to the annealing weight of the diversity reward, since the model begins to exploit the most promising strategy it has discovered. The final plateau reflects stabilization in training, implying the model has situated around effective solutions. We also present pairwise diversity of MUPO on validation set in Fig.~\ref{fig:diversity}~(A), which shows a much more gradual decline in contrast to sharp collapse observed in GRPO, indicating MUPO achieves a balanced trade-off between exploration and exploitation.

\begin{figure}[t!] 
    \centering
\includegraphics[width=0.96\linewidth]{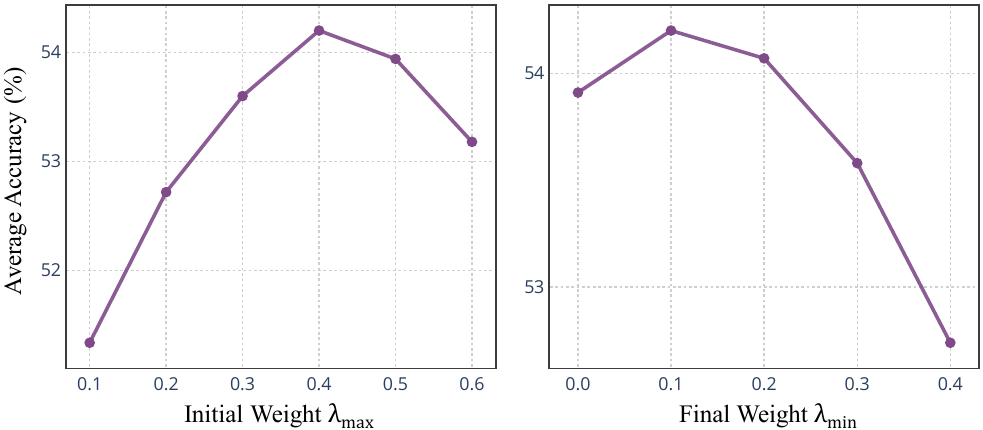}
    \vspace{-2mm}
    \caption{The ablation study of initial and final values of diversity reward weight $\lambda_{\rm max}$ and $\lambda_{\rm min}$ on average benchmark accuracy.
    }
\label{fig:weight}
\vspace{-4mm}
\end{figure}

\noindent\textbf{The number of groups $K$}. In Table~\ref{tab:ablation_k}, we investigate the impact of number of groups $K$ on accuracy. Notably, when $K = 1$, the training process degrades to GRPO. As $K$ increases, the accuracy rises rapidly and peaks at $K=3$. Moreover, for mathematical benchmarks, larger $K$ yields better performance, whereas smaller $K$ is more suitable for general problems. This reflects an intrinsic task nature: the former desires more flexible and diverse strategies, while the latter favors more uniform and structured reasoning.

\noindent\textbf{The diversity reward weight $\lambda_{\rm max}$ and $\lambda_{\rm \min}$}. MUPO adopts an annealing schedule for the diversity reward weight, progressively degrading from the initial value $\lambda_{\rm max}=0.4$ to the final value $\lambda_{\rm min}=0.1$ over the course of training. As shown in Fig.~\ref{fig:weight}, we perform an ablation study by fixing one of the parameters as its default while varying the other. The results show the optimal values align with our default settings. Increasing either parameter causes the diversity reward to dominate at the expense of accuracy, while reducing them weakens exploration, amplifying the risk of convergence to local optima. More results such as ablation of $\beta$ and limitation analysis are provided in Appendix.
\section{Conclusion}

In this paper, we identify a fundamental distinction between the behavioral patterns of RL and base models, where the former tends to engage in deeper yet narrow reasoning trajectories, while base models, despite less refined along individual path, exihibit broader and more diverse reasoning strategies. Through further analysis of training dynamics, we find that GRPO is prone to diversity collapse, causing models to converge to a limited set of strategies while neglecting the majority of potential alternatives, leading to local optima and poor scalability. To address this, we propose MUPO, a simple yet effective policy algorithm designed to incentivizes divergent thinking across multiple solutions and demonstrate its effectivenss on established benchmarks.

\noindent
\subsubsection*{Acknowledgement}
\label{sec:acknowledgement}
This research was, in part, funded by the U.S. Government – DARPA TIAMAT HR00112490421. The views and conclusions contained in this document are those of the authors and should not be interpreted as representing the official policies, either expressed or implied, of the U.S. Government.
{
    \small
    \bibliographystyle{ieeenat_fullname}
    \bibliography{main}

@String(CVPR= {IEEE Conf. Comput. Vis. Pattern Recog.})

@String(ECCV= {Eur. Conf. Comput. Vis.})

@String(ICLR = {Int. Conf. Learn. Represent.})

@String(CVPR  = {CVPR})

@String(ECCV  = {ECCV})

@String(ICLR  = {ICLR})

@article{kojima2022large,
  title={Large language models are zero-shot reasoners},
  author={Kojima, Takeshi and Gu, Shixiang Shane and Reid, Machel and Matsuo, Yutaka and Iwasawa, Yusuke},
  journal={NeurIPS},
  volume={35},
  pages={22199--22213},
  year={2022}
}

@article{muennighoff2025s1,
  title={s1: Simple test-time scaling},
  author={Muennighoff, Niklas and Yang, Zitong and Shi, Weijia and Li, Xiang Lisa and Fei-Fei, Li and Hajishirzi, Hannaneh and Zettlemoyer, Luke and Liang, Percy and Cand{\`e}s, Emmanuel and Hashimoto, Tatsunori},
  journal={arXiv preprint arXiv:2501.19393},
  year={2025}
}

@article{yang2025r1,
  title={R1-onevision: Advancing generalized multimodal reasoning through cross-modal formalization},
  author={Yang, Yi and He, Xiaoxuan and Pan, Hongkun and Jiang, Xiyan and Deng, Yan and Yang, Xingtao and Lu, Haoyu and Yin, Dacheng and Rao, Fengyun and Zhu, Minfeng and others},
  journal={arXiv preprint arXiv:2503.10615},
  year={2025}
}

@article{huang2025vision,
  title={Vision-r1: Incentivizing reasoning capability in multimodal large language models},
  author={Huang, Wenxuan and Jia, Bohan and Zhai, Zijie and Cao, Shaosheng and Ye, Zheyu and Zhao, Fei and Xu, Zhe and Hu, Yao and Lin, Shaohui},
  journal={arXiv preprint arXiv:2503.06749},
  year={2025}
}

@article{shao2024deepseekmath,
  title={Deepseekmath: Pushing the limits of mathematical reasoning in open language models},
  author={Shao, Zhihong and Wang, Peiyi and Zhu, Qihao and Xu, Runxin and Song, Junxiao and Bi, Xiao and Zhang, Haowei and Zhang, Mingchuan and Li, YK and Wu, Yang and others},
  journal={arXiv preprint arXiv:2402.03300},
  year={2024}
}

@article{liu2025visual,
  title={Visual-rft: Visual reinforcement fine-tuning},
  author={Liu, Ziyu and Sun, Zeyi and Zang, Yuhang and Dong, Xiaoyi and Cao, Yuhang and Duan, Haodong and Lin, Dahua and Wang, Jiaqi},
  journal={arXiv preprint arXiv:2503.01785},
  year={2025}
}

@article{shen2025vlm,
  title={Vlm-r1: A stable and generalizable r1-style large vision-language model},
  author={Shen, Haozhan and Liu, Peng and Li, Jingcheng and Fang, Chunxin and Ma, Yibo and Liao, Jiajia and Shen, Qiaoli and Zhang, Zilun and Zhao, Kangjia and Zhang, Qianqian and others},
  journal={arXiv preprint arXiv:2504.07615},
  year={2025}
}

@article{li2025think,
  title={Think or not think: A study of explicit thinking in rule-based visual reinforcement fine-tuning},
  author={Li, Ming and Zhong, Jike and Zhao, Shitian and Lai, Yuxiang and Zhang, Haoquan and Zhu, Wang Bill and Zhang, Kaipeng},
  journal={arXiv preprint arXiv:2503.16188},
  year={2025}
}

@article{xia2025visionary,
  title={Visionary-r1: Mitigating shortcuts in visual reasoning with reinforcement learning},
  author={Xia, Jiaer and Zang, Yuhang and Gao, Peng and Li, Yixuan and Zhou, Kaiyang},
  journal={arXiv preprint arXiv:2505.14677},
  year={2025}
}

@article{yang2025qwen3,
  title={Qwen3 technical report},
  author={Yang, An and Li, Anfeng and Yang, Baosong and Zhang, Beichen and Hui, Binyuan and Zheng, Bo and Yu, Bowen and Gao, Chang and Huang, Chengen and Lv, Chenxu and others},
  journal={arXiv preprint arXiv:2505.09388},
  year={2025}
}

@article{cai2024internlm2,
  title={Internlm2 technical report},
  author={Cai, Zheng and Cao, Maosong and Chen, Haojiong and Chen, Kai and Chen, Keyu and Chen, Xin and Chen, Xun and Chen, Zehui and Chen, Zhi and Chu, Pei and others},
  journal={arXiv preprint arXiv:2403.17297},
  year={2024}
}

@article{yu2025dapo,
  title={Dapo: An open-source llm reinforcement learning system at scale},
  author={Yu, Qiying and Zhang, Zheng and Zhu, Ruofei and Yuan, Yufeng and Zuo, Xiaochen and Yue, Yu and Dai, Weinan and Fan, Tiantian and Liu, Gaohong and Liu, Lingjun and others},
  journal={arXiv preprint arXiv:2503.14476},
  year={2025}
}

@article{snell2024scaling,
  title={Scaling llm test-time compute optimally can be more effective than scaling model parameters},
  author={Snell, Charlie and Lee, Jaehoon and Xu, Kelvin and Kumar, Aviral},
  journal={ICLR},
  year={2025}
}

@article{peng2025skywork,
  title={Skywork r1v: Pioneering multimodal reasoning with chain-of-thought},
  author={Peng, Yi and Wang, Peiyu and Wang, Xiaokun and Wei, Yichen and Pei, Jiangbo and Qiu, Weijie and Jian, Ai and Hao, Yunzhuo and Pan, Jiachun and Xie, Tianyidan and others},
  journal={arXiv preprint arXiv:2504.05599},
  year={2025}
}

@article{tan2025reason,
  title={Reason-rft: Reinforcement fine-tuning for visual reasoning},
  author={Tan, Huajie and Ji, Yuheng and Hao, Xiaoshuai and Lin, Minglan and Wang, Pengwei and Wang, Zhongyuan and Zhang, Shanghang},
  journal={arXiv preprint arXiv:2503.20752},
  year={2025}
}

@article{chen2025sft,
  title={Sft or rl? an early investigation into training r1-like reasoning large vision-language models},
  author={Chen, Hardy and Tu, Haoqin and Wang, Fali and Liu, Hui and Tang, Xianfeng and Du, Xinya and Zhou, Yuyin and Xie, Cihang},
  journal={arXiv preprint arXiv:2504.11468},
  year={2025}
}

@article{meng2025mm,
  title={Mm-eureka: Exploring the frontiers of multimodal reasoning with rule-based reinforcement learning},
  author={Meng, Fanqing and Du, Lingxiao and Liu, Zongkai and Zhou, Zhixiang and Lu, Quanfeng and Fu, Daocheng and Han, Tiancheng and Shi, Botian and Wang, Wenhai and He, Junjun and others},
  journal={arXiv preprint arXiv:2503.07365},
  year={2025}
}

@article{wang2025perception,
  title={Perception-aware policy optimization for multimodal reasoning},
  author={Wang, Zhenhailong and Guo, Xuehang and Stoica, Sofia and Xu, Haiyang and Wang, Hongru and Ha, Hyeonjeong and Chen, Xiusi and Chen, Yangyi and Yan, Ming and Huang, Fei and others},
  journal={arXiv preprint arXiv:2507.06448},
  year={2025}
}

@inproceedings{zhang2024mathverse,
  title={Mathverse: Does your multi-modal llm truly see the diagrams in visual math problems?},
  author={Zhang, Renrui and Jiang, Dongzhi and Zhang, Yichi and Lin, Haokun and Guo, Ziyu and Qiu, Pengshuo and Zhou, Aojun and Lu, Pan and Chang, Kai-Wei and Qiao, Yu and others},
  booktitle={ECCV},
  pages={169--186},
  year={2024},
  organization={Springer}
}

@article{xiao2024logicvista,
  title={Logicvista: Multimodal llm logical reasoning benchmark in visual contexts},
  author={Xiao, Yijia and Sun, Edward and Liu, Tianyu and Wang, Wei},
  journal={arXiv preprint arXiv:2407.04973},
  year={2024}
}

@article{chen2024we,
  title={Are we on the right way for evaluating large vision-language models?},
  author={Chen, Lin and Li, Jinsong and Dong, Xiaoyi and Zhang, Pan and Zang, Yuhang and Chen, Zehui and Duan, Haodong and Wang, Jiaqi and Qiao, Yu and Lin, Dahua and others},
  journal={NeurIPS},
  volume={37},
  pages={27056--27087},
  year={2024}
}

@inproceedings{guan2024hallusionbench,
  title={Hallusionbench: an advanced diagnostic suite for entangled language hallucination and visual illusion in large vision-language models},
  author={Guan, Tianrui and Liu, Fuxiao and Wu, Xiyang and Xian, Ruiqi and Li, Zongxia and Liu, Xiaoyu and Wang, Xijun and Chen, Lichang and Huang, Furong and Yacoob, Yaser and others},
  booktitle={CVPR},
  pages={14375--14385},
  year={2024}
}

@misc{Singh_2025, 
    title={GPT-5 system card}, 
    url={https://openai.com/index/gpt-5-system-card/}, 
    journal={OpenAI}, 
    author={Singh, Aaditya and Fry, Adam and Perelman, Adam and others}, 
    year={2025}}

@article{yang2025look,
  title={Look-Back: Implicit Visual Re-focusing in MLLM Reasoning},
  author={Yang, Shuo and Niu, Yuwei and Liu, Yuyang and Ye, Yang and Lin, Bin and Yuan, Li},
  journal={arXiv preprint arXiv:2507.03019},
  year={2025}
}

@article{vl-rethinker,
      title={VL-Rethinker: Incentivizing Self-Reflection of Vision-Language Models with Reinforcement Learning},
      author = {Wang, Haozhe and Qu, Chao and Huang, Zuming and Chu, Wei and Lin,Fangzhen and Chen, Wenhu},
      journal={arXiv preprint arXiv:2504.08837},
      year={2025}
}

@article{bai2025qwen2,
  title={Qwen2. 5-vl technical report},
  author={Bai, Shuai and Chen, Keqin and Liu, Xuejing and Wang, Jialin and Ge, Wenbin and Song, Sibo and Dang, Kai and Wang, Peng and Wang, Shijie and Tang, Jun and others},
  journal={arXiv preprint arXiv:2502.13923},
  year={2025}
}

@article{lu2023mathvista,
  title={Mathvista: Evaluating mathematical reasoning of foundation models in visual contexts},
  author={Lu, Pan and Bansal, Hritik and Xia, Tony and Liu, Jiacheng and Li, Chunyuan and Hajishirzi, Hannaneh and Cheng, Hao and Chang, Kai-Wei and Galley, Michel and Gao, Jianfeng},
  journal={ICLR},
  year={2023}
}

@article{wang2024measuring,
  title={Measuring multimodal mathematical reasoning with math-vision dataset},
  author={Wang, Ke and Pan, Junting and Shi, Weikang and Lu, Zimu and Ren, Houxing and Zhou, Aojun and Zhan, Mingjie and Li, Hongsheng},
  journal={NeurIPS},
  volume={37},
  pages={95095--95169},
  year={2024}
}

@article{qiao2024we,
  title={We-math: Does your large multimodal model achieve human-like mathematical reasoning?},
  author={Qiao, Runqi and Tan, Qiuna and Dong, Guanting and Wu, Minhui and Sun, Chong and Song, Xiaoshuai and GongQue, Zhuoma and Lei, Shanglin and Wei, Zhe and Zhang, Miaoxuan and others},
  journal={ACL},
  year={2024}
}

@article{lu2021inter,
  title={Inter-gps: Interpretable geometry problem solving with formal language and symbolic reasoning},
  author={Lu, Pan and Gong, Ran and Jiang, Shibiao and Qiu, Liang and Huang, Siyuan and Liang, Xiaodan and Zhu, Song-Chun},
  journal={ACL},
  year={2021}
}

@article{yu2023mm,
  title={Mm-vet: Evaluating large multimodal models for integrated capabilities},
  author={Yu, Weihao and Yang, Zhengyuan and Li, Linjie and Wang, Jianfeng and Lin, Kevin and Liu, Zicheng and Wang, Xinchao and Wang, Lijuan},
  journal={ICML},
  year={2023}
}

@article{comanici2025gemini,
  title={Gemini 2.5: Pushing the frontier with advanced reasoning, multimodality, long context, and next generation agentic capabilities},
  author={Comanici, Gheorghe and Bieber, Eric and Schaekermann, Mike and Pasupat, Ice and Sachdeva, Noveen and Dhillon, Inderjit and Blistein, Marcel and Ram, Ori and Zhang, Dan and Rosen, Evan and others},
  journal={arXiv preprint arXiv:2507.06261},
  year={2025}
}

@article{chen2024expanding,
  title={Expanding performance boundaries of open-source multimodal models with model, data, and test-time scaling},
  author={Chen, Zhe and Wang, Weiyun and Cao, Yue and Liu, Yangzhou and Gao, Zhangwei and Cui, Erfei and Zhu, Jinguo and Ye, Shenglong and Tian, Hao and Liu, Zhaoyang and others},
  journal={arXiv preprint arXiv:2412.05271},
  year={2024}
}

@article{touvron2023llama,
  title={Llama: Open and efficient foundation language models},
  author={Touvron, Hugo and Lavril, Thibaut and Izacard, Gautier and Martinet, Xavier and Lachaux, Marie-Anne and Lacroix, Timoth{\'e}e and Rozi{\`e}re, Baptiste and Goyal, Naman and Hambro, Eric and Azhar, Faisal and others},
  journal={arXiv preprint arXiv:2302.13971},
  year={2023}
}

@article{schulman2017proximal,
  title={Proximal policy optimization algorithms},
  author={Schulman, John and Wolski, Filip and Dhariwal, Prafulla and Radford, Alec and Klimov, Oleg},
  journal={arXiv preprint arXiv:1707.06347},
  year={2017}
}

@article{zhou2022least,
  title={Least-to-most prompting enables complex reasoning in large language models},
  author={Zhou, Denny and Sch{\"a}rli, Nathanael and Hou, Le and Wei, Jason and Scales, Nathan and Wang, Xuezhi and Schuurmans, Dale and Cui, Claire and Bousquet, Olivier and Le, Quoc and others},
  journal={ICLR},
  year={2022}
}

@article{yue2025does,
  title={Does reinforcement learning really incentivize reasoning capacity in llms beyond the base model?},
  author={Yue, Yang and Chen, Zhiqi and Lu, Rui and Zhao, Andrew and Wang, Zhaokai and Song, Shiji and Huang, Gao},
  journal={arXiv preprint arXiv:2504.13837},
  year={2025}
}

@article{zhang2025gvpo,
  title={GVPO: Group variance policy optimization for large language model post-training},
  author={Zhang, Kaichen and Hong, Yuzhong and Bao, Junwei and Jiang, Hongfei and Song, Yang and Hong, Dingqian and Xiong, Hui},
  journal={arXiv preprint arXiv:2504.19599},
  year={2025}
}

@article{tian2025more,
  title={More thought, less accuracy? on the dual nature of reasoning in vision-language models},
  author={Tian, Xinyu and Zou, Shu and Yang, Zhaoyuan and He, Mengqi and Waschkowski, Fabian and Wesemann, Lukas and Tu, Peter and Zhang, Jing},
  journal={arXiv preprint arXiv:2509.25848},
  year={2025}
}

@article{jian2025look,
  title={Look again, think slowly: Enhancing visual reflection in vision-language models},
  author={Jian, Pu and Wu, Junhong and Sun, Wei and Wang, Chen and Ren, Shuo and Zhang, Jiajun},
  journal={arXiv preprint arXiv:2509.12132},
  year={2025}
}

@article{wang2025beyond,
  title={Beyond the 80/20 rule: High-entropy minority tokens drive effective reinforcement learning for llm reasoning},
  author={Wang, Shenzhi and Yu, Le and Gao, Chang and Zheng, Chujie and Liu, Shixuan and Lu, Rui and Dang, Kai and Chen, Xionghui and Yang, Jianxin and Zhang, Zhenru and others},
  journal={arXiv preprint arXiv:2506.01939},
  year={2025}
}

@article{cheng2025reasoning,
  title={Reasoning with exploration: An entropy perspective},
  author={Cheng, Daixuan and Huang, Shaohan and Zhu, Xuekai and Dai, Bo and Zhao, Wayne Xin and Zhang, Zhenliang and Wei, Furu},
  journal={arXiv preprint arXiv:2506.14758},
  year={2025}
}

@article{zheng2025parallel,
  title={Parallel-r1: Towards parallel thinking via reinforcement learning},
  author={Zheng, Tong and Zhang, Hongming and Yu, Wenhao and Wang, Xiaoyang and Yang, Xinyu and Dai, Runpeng and Liu, Rui and Bao, Huiwen and Huang, Chengsong and Huang, Heng and others},
  journal={arXiv preprint arXiv:2509.07980},
  year={2025}
}

@article{wang2025visualprm,
  title={Visualprm: An effective process reward model for multimodal reasoning},
  author={Wang, Weiyun and Gao, Zhangwei and Chen, Lianjie and Chen, Zhe and Zhu, Jinguo and Zhao, Xiangyu and Liu, Yangzhou and Cao, Yue and Ye, Shenglong and Zhu, Xizhou and others},
  journal={arXiv preprint arXiv:2503.10291},
  year={2025}
}

@article{zhang2025qwen3,
  title={Qwen3 Embedding: Advancing Text Embedding and Reranking Through Foundation Models},
  author={Zhang, Yanzhao and Li, Mingxin and Long, Dingkun and Zhang, Xin and Lin, Huan and Yang, Baosong and Xie, Pengjun and Yang, An and Liu, Dayiheng and Lin, Junyang and others},
  journal={arXiv preprint arXiv:2506.05176},
  year={2025}
}

@article{peng2025lmm,
  title={Lmm-r1: Empowering 3b lmms with strong reasoning abilities through two-stage rule-based rl},
  author={Peng, Yingzhe and Zhang, Gongrui and Zhang, Miaosen and You, Zhiyuan and Liu, Jie and Zhu, Qipeng and Yang, Kai and Xu, Xingzhong and Geng, Xin and Yang, Xu},
  journal={arXiv preprint arXiv:2503.07536},
  year={2025}
}

@article{jiang2025risk,
  title={Risk-Sensitive RL for Alleviating Exploration Dilemmas in Large Language Models},
  author={Jiang, Yuhua and Huang, Jiawei and Yuan, Yufeng and Mao, Xin and Yue, Yu and Zhao, Qianchuan and Yan, Lin},
  journal={arXiv preprint arXiv:2509.24261},
  year={2025}
}

@article{bradley2000constrained,
  title={Constrained k-means clustering},
  author={Bradley, Paul S and Bennett, Kristin P and Demiriz, Ayhan},
  journal={Microsoft Research, Redmond},
  volume={20},
  number={0},
  pages={0},
  year={2000}
}

@article{seed2025seed1,
  title={Seed1. 5-thinking: Advancing superb reasoning models with reinforcement learning},
  author={Seed, ByteDance and Chen, Jiaze and Fan, Tiantian and Liu, Xin and Liu, Lingjun and Lin, Zhiqi and Wang, Mingxuan and Wang, Chengyi and Wei, Xiangpeng and Xu, Wenyuan and others},
  journal={arXiv preprint arXiv:2504.13914},
  year={2025}
}

@article{guo2025seed1,
  title={Seed1. 5-vl technical report},
  author={Guo, Dong and Wu, Faming and Zhu, Feida and Leng, Fuxing and Shi, Guang and Chen, Haobin and Fan, Haoqi and Wang, Jian and Jiang, Jianyu and Wang, Jiawei and others},
  journal={arXiv preprint arXiv:2505.07062},
  year={2025}
}

@article{hu2025diversity,
  title={Diversity-Incentivized Exploration for Versatile Reasoning},
  author={Hu, Zican and Zhang, Shilin and Li, Yafu and Yan, Jianhao and Hu, Xuyang and Cui, Leyang and Qu, Xiaoye and Chen, Chunlin and Cheng, Yu and Wang, Zhi},
  journal={arXiv preprint arXiv:2509.26209},
  year={2025}
}

@inproceedings{tian2024argue,
  title={Argue: Attribute-guided prompt tuning for vision-language models},
  author={Tian, Xinyu and Zou, Shu and Yang, Zhaoyuan and Zhang, Jing},
  booktitle={Proceedings of the IEEE/CVF Conference on Computer Vision and Pattern Recognition},
  pages={28578--28587},
  year={2024}
}

@inproceedings{tian2025identifying,
  title={Identifying and mitigating position bias of multi-image vision-language models},
  author={Tian, Xinyu and Zou, Shu and Yang, Zhaoyuan and Zhang, Jing},
  booktitle={Proceedings of the Computer Vision and Pattern Recognition Conference},
  pages={10599--10609},
  year={2025}
}

@article{yao2023training,
  title={Training with product digital twins for autoretail checkout},
  author={Yao, Yue and Tian, Xinyu and Tang, Zheng and Biswas, Sujit and Lei, Huan and Gedeon, Tom and Zheng, Liang},
  journal={arXiv preprint arXiv:2308.09708},
  year={2023}
}

@inproceedings{zou2025simlabel,
  title={Simlabel: Consistency-guided ood detection with pretrained vision-language models},
  author={Zou, Shu and Tian, Xinyu and Zhao, Qinyu and Yang, Zhaoyuan and Zhang, Jing},
  booktitle={Australasian Joint Conference on Artificial Intelligence},
  pages={110--121},
  year={2025},
  organization={Springer}
}

@article{tian2025black,
  title={Black sheep in the herd: Playing with spuriously correlated attributes for vision-language recognition},
  author={Tian, Xinyu and Zou, Shu and Yang, Zhaoyuan and He, Mengqi and Zhang, Jing},
  journal={arXiv preprint arXiv:2502.15809},
  year={2025}
}

@inproceedings{zou2026unlocking,
  title={Unlocking vision-language models for video anomaly detection via fine-grained prompting},
  author={Zou, Shu and Tian, Xinyu and Wesemann, Lukas and Waschkowski, Fabian and Yang, Zhaoyuan and Zhang, Jing},
  booktitle={Proceedings of the IEEE/CVF Winter Conference on Applications of Computer Vision},
  pages={4223--4233},
  year={2026}
}

@article{yao2025simple,
  title={Simple radiology vllm test-time scaling with thought graph traversal},
  author={Yao, Yue and Wen, Zelin and Tong, Yan and Tian, Xinyu and Li, Xuqing and Ma, Xiao and Xu, Dongliang and Gedeon, Tom},
  journal={arXiv preprint arXiv:2506.11989},
  year={2025}
}

@article{he2025few,
  title={Few Tokens Matter: Entropy Guided Attacks on Vision-Language Models},
  author={He, Mengqi and Tian, Xinyu and Shen, Xin and Ni, Jinhong and Zou, Shu and Yang, Zhaoyuan and Zhang, Jing},
  journal={arXiv preprint arXiv:2512.21815},
  year={2025}
}
}


\end{document}